\documentclass[conference,9pt]{IEEEtran}
\usepackage{cite}
\usepackage{amsmath,amssymb,amsfonts}
\usepackage{booktabs}
\usepackage{colortbl}
\usepackage{graphicx}
\usepackage{ifthen}
\usepackage{subcaption}
\usepackage{xcolor}
\usepackage{tcolorbox}
\usepackage{hyperref}
\usepackage{adjustbox}
\usepackage{enumitem}

\newboolean{anonymous}
\setboolean{anonymous}{false} 

\newcommand{\schemato}{\texorpdfstring{$Schemato$}{Schemato}}

\newcommand{\llama}{Llama-3.1-8B}

\colorlet{tablerowcolor}{gray!10} 
\newcommand{\rowcol}{\rowcolor{tablerowcolor}} %

\begin{document}

\title{\schemato{} -- An LLM for Netlist-to-Schematic Conversion}

\ifthenelse{\boolean{anonymous}}{
    \author{
        \IEEEauthorblockN{Anonymous Authors}
        \IEEEauthorblockA{Anonymous Affiliations}}
}{
    \author{
        \IEEEauthorblockN{Ryoga Matsuo\textsuperscript{1,2}, Stefan Uhlich\textsuperscript{3}, Arun Venkitaraman\textsuperscript{1}, Andrea Bonetti\textsuperscript{1}, Chia-Yu Hsieh\textsuperscript{1}, Ali Momeni\textsuperscript{1,2}}
        \IEEEauthorblockN{Lukas Mauch\textsuperscript{3}, Augusto Capone\textsuperscript{3}, Eisaku Ohbuchi\textsuperscript{4}, Lorenzo Servadei\textsuperscript{1,5}}
        \IEEEauthorblockA{\textsuperscript{1}\textit{SonyAI, Switzerland}\quad \textsuperscript{2}\textit{EPFL, Switzerland}}
        \IEEEauthorblockA{\textsuperscript{3}\textit{Sony Semiconductor Solutions Europe, Germany} \quad\textsuperscript{4}\textit{Sony Semiconductor Solutions, Japan} \quad \textsuperscript{5}\textit{TU Munich, Germany}}}
}


\maketitle

\begin{abstract}
Machine learning models are advancing circuit design, particularly in analog circuits. They typically generate netlists that lack human interpretability. This is a problem as human designers heavily rely on the interpretability of circuit diagrams or schematics to intuitively understand, troubleshoot, and develop designs. Hence, to integrate domain knowledge effectively, it is crucial to translate ML-generated netlists into interpretable schematics quickly and accurately. We propose Schemato, a large language model (LLM) for netlist-to-schematic conversion. In particular, we consider our approach in converting netlists to .asc files, text-based schematic description used in LTSpice. Experiments on our circuit dataset show that Schemato achieves up to 76\% compilation success rate, surpassing 63\% scored by the state-of-the-art LLMs. Furthermore, our experiments show that Schemato generates schematics with an average graph edit distance score and mean structural similarity index measure, scaled by the compilation success rate that are 1.8$\times$ and 4.3$\times$ higher than the best performing LLMs respectively, demonstrating its ability to generate schematics that are more accurately connected and are closer to the reference human design.
\end{abstract}


\section{Introduction}
\label{sec:intro}

Recent years have seen a rapid increase in the number of machine learning (ML)-based solutions that address different aspects of analog design, specially towards both topology and sizing \cite{GANA,DomainAdaptedRL,ALIGN,Yin2024ADOLLMAD,cktcompletion,uhlich2024graco}. The netlist representation of the analog circuit forms a central part of all these approaches, since they all operate by either encoding the information in the netlist or generating a netlist as part of an end-to-end ML model. However, often such circuits, while they may maximize a certain objective function and, hence, are feasible in the eyes of an ML practician, might not result in a practical design in the eyes of an analog designer. An analog designer usually views and assesses a circuit through the circuit diagram or schematics and not a netlist. This means that having a schematic or circuit diagram corresponding to a netlist is crucial for the ML models engineer to benefit from the domain knowledge of the designer's feedback.  This is specially the case when the circuits become large in size and when one considers problems such as synthesizing circuits \cite{DBLP:conf/date/SettaluriH0HN20,DBLP:journals/tcad/ZhaoZ22}, where the designer must spend considerable resources in drawing a schematic by hand each time to be able to give feedback to the ML models more effectively. Thus, it is highly desirable to be able to efficiently and automatically translate netlists into circuit diagrams or schematics. 

\begin{figure*}
\centering
    \includegraphics[width=1.0\linewidth]{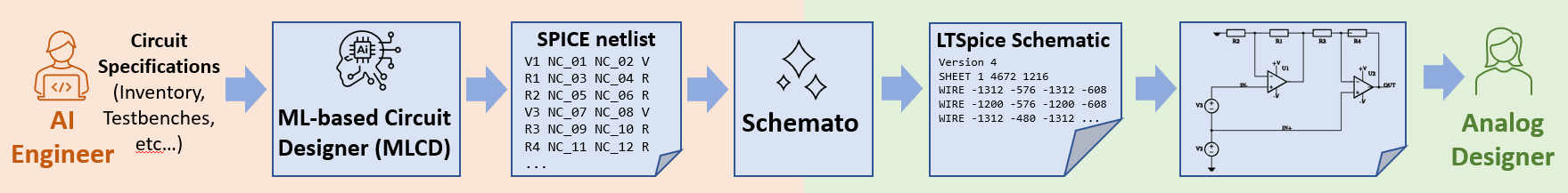}
    \caption{Use case of \schemato{}: AI conversion of SPICE netlists into human-readable schematics.}
    \label{fig_schemato}
    \vspace{-0.3cm}
\end{figure*}

Traditionally, circuit schematics are manually generated using conventional electronic design automation (EDA) tools by analog designers and visualized with associated viewing tools, such as LTSpice, xschem and Cadence Virtuoso. LTSpice and xschem process text-based structured schematic files (.asc for LTSpice and .sch for xschem) to produce circuit diagrams \cite{LTSpiceSimulator, xschem}. These methods rely on structured programs with specific semantics or ``languages".
Similarly, netlists adhere to well-defined semantics for circuit representation, making the task of conversion to a schematic or diagram analogous to a language translation task. This analogy highlights the potential of leveraging large language models (LLMs), which excel at understanding and predicting semantic structures \cite{openai2024gpt4technicalreport,dubey2024llama3herdmodels}. LLMs have also recently shown great potential in different aspects of EDA, particularly, in digital and logic synthesis \cite{llm4eda1,llm4eda2,llm4eda3,llm4eda4,Yin2024ADOLLMAD,lai2024analogcoder} wherein the models learn to understand, abstract, and build upon the design process of human designers. This motivates us to propose an LLM-based solution for translating netlists to circuit schematics/diagrams. 

The earliest works on automated netlist to schematic generation have mostly been in the domain of digital ICs \cite{naveen1993automatic,lageweg1998designing}. Some of the early works for analog circuits include \cite{arsintescu1996analog} that introduced a symmetry-based placement algorithm that identifies symmetrical structures within a netlist, and the work of \cite{lee1992aesthetic} that concentrated on schematic routing aimed at minimizing net crossings.
The more recent work of \cite{hsu2022automatic} presents a ML-based approach that combines identification of subblocks with an RL block that generates the schematic by optimizing for a reward based on the building block compliance rate that measures aesthetic of the generated schematic based on topological and heuristic constraints.

While these approaches have been shown to perform well on different circuit topologies, they rely on either the explicit use of heuristic rules or an understanding of the underlying topology/hierarchy of its subblocks or subcircuits or on the use of specific aesthetic reward functions. \schemato{} overcomes these limitations by developing an automated netlist to schematic conversion system that is independent of the circuit type or functionality, enabling broad applicability across diverse designs without requiring prior knowledge of specific circuit characteristics, while maintaining as faithfully as possible the aesthetics of a human design. This is because our approach learns from human-designed circuit examples and therefore does not require an explicit aesthetics measure. We also note that due to the universal nature of the framework, we are not limited to specific schematic templates or the choice of the schematic editor; although we have used LTSpice with the .asc file format in our experiments, our approach is equally valid with the use of other EDA tools with their schematic formats.

\noindent In particular, our contributions are:
\begin{itemize}
    \item We propose \schemato{}, an LLM for automatic translation of netlists to text-based schematic representations, fine-tuned on human-created design examples.
    \item In particular, we perform a task of netlist-to-schematic conversion, namely, netlist-to-asc conversion for use with LTSpice.
    \item We perform exhaustive experiments on publicly available circuit and evaluate the performance of our approach using various metrics such as graph edit distance (GED) scores, mean structural similarity index measure (MSSIM), compilation success rate (CSR), and bilingual evaluation understudy (BLEU).
    \item Experiments show that \schemato{} outperforms state-of-the-art LLMs with a CSR of 76\%, a 1.8$\times$ higher average GED score scaled by CSR, and a 4.3$\times$ higher average MSSIM scaled by CSR over the test set.
\end{itemize}
To the best of our knowledge, no prior work exists that addresses this specific and important problem using LLMs, and given the increasing influence of LLMs, we believe our contribution is both timely and valuable for advancing ML-based approaches for analog design. 

\section{Preliminaries}

In this section, we present a short review of LLMs followed by a short review of the LTSpice .asc schematic format that builds the basis of our current work.

\subsection{Review of LLMs}
Large Language Models (LLMs) are advanced AI systems capable of performing complex linguistic tasks such as translation, summarization, and reasoning by leveraging large datasets and neural architectures \cite{openai2024gpt4technicalreport,dubey2024llama3herdmodels}. Their effectiveness often hinges on prompt engineering, with techniques like Chain-of-Thought and few-shot prompting enabling more accurate and nuanced outputs \cite{chainofthought}. However, as most LLMs are trained on general-purpose data, their performance in specialized domains like EDA may be limited due to the lack of task-specific optimization \cite{llm4eda1,llm4eda2,llm4eda3,lai2024analogcoder}. Addressing this gap typically requires fine-tuning or domain-specific prompt strategies.


In our work, we consider different combinations of prompt design, in-context examples, and fine-tuning of LLMs to translate netlists into circuit schematics.  Our experiments demonstrate that while both prompt engineering and the incorporation of few-shot examples are crucial for improving the output of the original LLM, fine-tuning yields a substantially enhanced LLM, called {\schemato{}}.

\subsection{Review of LTSpice-specific .asc file format} \label{Ch:asc_description}
\label{sec:ltspice_review}
We next shortly review the key terms in the LTSpice .asc file\footnote{More details may be found in the documentation of LTSpice \cite{LTSpiceSimulator}.} As we discuss later, this will also motivate some of the prompts used in our experiments. In the .asc file, each line starts with one of the following keywords accompanied by the corresponding input arguments 
\begin{itemize}
    \item ``\texttt{Version index}'' defines the LTSpice version.
    \item ``\texttt{SHEET index height width}'': defines the sheet index and size on the LTSpice application.
    \item ``\texttt{WIRE $x_{start}$ $y_{start}$ $x_{end}$ $y_{end}$}": creates a wire that extends from position ($x_{start}$, $y_{start}$) to ($x_{end}$, $y_{end}$).
    \item ``\texttt{SYMBOL component $x_{comp}$ $y_{comp}$ $\theta$}'': instantiates a circuit component (e.g. nmos, pmos, resistor, opamp, inverter) at the specified position  ($x_{comp}$, $y_{comp}$) and orientation $\theta$, where $\theta=R90$ corresponds to $90^{\circ}$ anti-clockwise rotation and $\theta=M270$ for $270^{\circ}$ anti-clockwise rotation combined with mirroring. This line is followed by one of these lines that sets a symbol attribute:
    \begin{itemize}
        \item ``\texttt{SYMATTR InstName component\_name}'': assigns the component's name.
        \item ``\texttt{SYMATTR SpiceModel model\_name}'': assigns the component's spice model.
        \item ``\texttt{SYMATTR ModelFile model\_file}'': assigns the component's spice model path.
    \end{itemize}
    \end{itemize}

\begin{figure*}

    \begin{tcolorbox}[colback=gray!5!white, colframe=gray!75!black, title={\makebox[0.54\linewidth][l]{\scriptsize\textbf{Zero-shot}} 
    \makebox[0.4\linewidth][l]{\scriptsize\textbf{One-shot}}}]
    \begin{minipage}{0.5\textwidth} 
        \vspace{-0.2cm}
        \scriptsize
        \textbf{Prompt 1: Simply instruct the LLM for the task}\\
        Convert the following netlist to a .asc file, a text-based circuit schematic file for LTSpice. Only output the code and contain it in ```.

        \medskip
        \textbf{Input:} ```\texttt{Netlist}```

        ---

        \textbf{Output:}

        \hrulefill

        \textbf{Prompt 2: Specify the types of keywords to start each line of the output}\\
        Convert the following netlist to a .asc file, a text-based circuit schematic file for LTSpice. Start each line with keywords such as `\texttt{Version}`, `\texttt{SHEET}`, `\texttt{WIRE}`, `\texttt{FLAG}`, `\texttt{SYMBOL}`, and `\texttt{SYMATTR}`. Only output the code and contain it in ```.

        \medskip
        \textbf{Input:} ```\texttt{Netlist}```

        ---

        \textbf{Output:}

        \hrulefill

        \textbf{Prompt 3: Specify the first two lines including the sheet size}\\
        Convert the following netlist to a .asc file, a text-based circuit schematic file for LTSpice. Start your answer with \\
        ```\texttt{Version 4 \\
        SHEET index height width} ```\\. Only output the code and contain it in ```.

        \medskip
        \textbf{Input:} ```\texttt{Netlist}```

        ---

        \textbf{Output:}
        
    \end{minipage}%
    \hfill
    \begin{minipage}{0.45\textwidth} 
        \vspace{-0.2cm}
        \scriptsize
        \textbf{Prompt 4: Provide an in-context example of conversion}\\
        Convert the following netlist to a .asc file, a text-based circuit schematic file for LTSpice. Only output the code and contain it in ```. \\
        
        Here is an example: 

        \textbf{Input:} ```Example netlist of a low-pass filter (LPF)```

        ---

        \textbf{Output:} ```Example .asc file of the LPF```

        \medskip
        \textbf{Input:} ```\texttt{Netlist}```

        ---

        \textbf{Output:}

        \hrulefill

        \textbf{Prompt 5: Combination of Prompt 3 and 4}\\
        Convert the following netlist to a .asc file, a text-based circuit schematic file for LTSpice. Start your answer with \\
        ```\texttt{Version 4 \\
        SHEET index height width} ```\\. Only output the code and contain it in ```. \\
        
        Here is an example: 

        \textbf{Input:} ```Example netlist of an LPF```

        ---

        \textbf{Output:} ```Example .asc file of the LPF```

        \medskip
        \textbf{Input:} ```\texttt{Netlist}```

        ---

        \textbf{Output:}
    \end{minipage}
\end{tcolorbox}
\vspace{-0.2cm}
\caption{Zero-shot \mbox{Prompt 1-3} and one-shot \mbox{Prompt 4-5} for netlist-to-schematic conversion.}
\label{fig:prompt1-5}
\vspace{-0.4cm}
\end{figure*}

\section{Proposed approach: \schemato{}}
In \autoref{fig_schemato}, we present an illustration of the \schemato{}-based circuit design workflow. The left half illustrates the fully data-driven model, while the right half integrates domain knowledge contributions from analog designers.

As discussed in \autoref{sec:intro}, \schemato{} addresses the task of automating circuit schematic generation from netlists. At its core, \schemato{} is an LLM fine-tuned using a curated dataset of human-created netlist-to-schematic pairs. Its development focuses on three key components:
\begin{itemize}
\item Task Definition: Constructing prompts that define the netlist-to-schematic conversion task. 
\item Guidance via Context: Incorporating in-context examples within prompts to guide the model. 
\item Fine-Tuning: Adapting the model using a specialized dataset of netlist-schematic pairs. \end{itemize}
\schemato{} supports the .asc file format for schematic generation. This file format is compatible with the LTSpice schematic viewer/editing tool. While our experiments focus on LTSpice due to its data availability, the approach generalizes to any text-based schematic format.

\subsection{Task definition and guidance via context}
We consider the task to translate netlists to LTSpice text-based schematics for the evaluation of different models. 
Furthermore, to develop and refine \schemato{}, we consider different prompting techniques to guide the LLM effectively. The summary of these prompt variants are provided in \autoref{fig:prompt1-5}.
As shown in the figure, \mbox{Prompt 1} is a simple, task-oriented description that serves as the baseline for comparison. \mbox{Prompt 2} introduces syntax-specific keywords relevant to the target format, such as LTSpice symbols(\autoref{sec:ltspice_review}).
\mbox{Prompt 3} enhances guidance by providing the initial two lines of the expected output, including the sheet size of the reference schematic. This guides the LLM to generate a schematic with the intended layout and representation. 
\mbox{Prompt 4} incorporate specific netlist-to-schematic conversion examples, offering the model concrete patterns for improved accuracy. Finally, \mbox{Prompt 5} is merely a combination of \mbox{Prompt 3} and 4. In summary, \mbox{Prompts 1-3} utilize zero-shot learning, while \mbox{Prompts 4-5} employ one-shot learning.

\subsection{Fine-tuning}

\schemato{} is developed by fine-tuning state-of-the-art LLMs using human-generated netlist-schematic data pairs. To address the limited availability of such data, we employ data augmentation techniques to expand the dataset, ensuring better coverage of diverse scenarios. To select the optimal base model, we first evaluate various LLMs using all proposed prompts. The combination of prompt and model yielding the highest performance is then fine-tuned resulting in \schemato{}.
The comprehensive details on fine-tuning methodology, data preprocessing, and augmentation strategies are provided in the next section.

\section{Experimental setup} \label{Ch:inference}
In our experiments, we use LTSpice schematics from GitHub, which contain circuits with diverse features and applications. We also include the publicly available LTSpice circuit files packaged with LTSpice XVII (examples in folders, Eduational \& Applications \cite{LTSpiceSimulator}). These sets are used for training and validation. In addition, we consider the LTSpice schematics contained in the public GitHub repository: {\em Circuits-LTSpice}\footnote{\url{https://github.com/mick001/Circuits-LTSpice}} for testing. The circuit examples in this repository are mostly composed of generic components such as resistors, capacitors, inductors, bipolar transistors, MOS transistors, voltage, and current sources. This feature of the test set facilitates the model validation for generalization. 
In all the cases, we start from the available .asc files. These are then processed through LTSpice to create the corresponding netlists. Some .asc files cannot be compiled into netlist due to missing symbols in the default libraries of LTSpice. These .asc files are omitted from our dataset. To ensure the validity of our method in measuring the generalization performance, we remove the training samples of .asc and netlist files that overlap with the existing test samples.
We next describe the details of data preprocessing and filtering.

\subsection{Preprocessing and filtering} 
\label{ch:preprocessing}
To facilitate {\em Schemato} to learn the translation task, we preprocess the dataset by removing unnecessary information for the schematic visualizations in LTSpice. This includes spice simulation commands, spice model definitions, user annotations, comments, etc. We explain how we preprocess .asc and netlist files below.

\noindent \textbf{LTSpice-specific .asc files}:

\begin{itemize}
    \item Modify the schematic sheet size defined by ``\texttt{SHEET index height width}'' by setting the height and width to the maximum vertical/horizontal coordinate minus the minimum vertical/horizontal coordinate out of all the components. \schemato{} can use the modified values to learn the geometries of the schematics.
    \item Modify the coordinates of all the wires and components such that the center of the schematic is defined at the origin $(0,0)$. This makes the dataset shift-invariant.

    \item 
Replace \texttt{SpiceModel} and \texttt{ModelFile} with \texttt{InstName} while using the original \texttt{model\_name} and \texttt{model\_file} for \texttt{component\_name}. This allows for any line with \texttt{SYMATTR} to be compiled and \texttt{component\_name} can be set arbitrarily.

\item Remove the lines starting with \textsuperscript{\texttt{*}}, \texttt{TEXT}, \texttt{RECTANGLE}, \texttt{WINDOW}, \texttt{LINE}, and \texttt{CIRCLE}. 
\item Remove the user annotations/comments since they are not needed for the schematic visualization by LTSpice. 
\item Discard .asc files that do not contain \texttt{SYMBOL}-\texttt{SYMATTR} pair as this type of example is not composed of meaningful circuit components except for wires.
\end{itemize}

\noindent \textbf{Netlists}:
The file format we use for netlist is .net.
We remove the lines with commands such as \texttt{ .backanno}, \texttt{ .lib}, and \texttt{.model} as they are not required for the schematic visualizations.

We next consider the augmentation strategy used to increase the effective dataset size.
\subsection{Data augmentation}
For \schemato{} to be agnostic to the line ordering of .asc files, we create an augmented data example by swapping the lines of the .asc files. Specifically, .asc files can have different orders of \texttt{SYMBOL}-\texttt{SYMATTR} pairs resulting in the same schematic representations. Therefore, we randomly shuffle their line ordering and generate five different augmentations for .asc files with more than three \texttt{SYMBOL}-\texttt{SYMATTR} pairs. For .asc files with two \texttt{SYMBOL}-\texttt{SYMATTR} pairs, we swap their order to generate two samples and we leave the files with one \texttt{SYMBOL}-\texttt{SYMATTR} pair unmodified as we cannot augment such samples.
After the generation of these augmented .asc file samples, we convert them into netlists to create the dataset as described in \autoref{ch:preprocessing}. The changes of the line ordering in .asc files are reflected in the line ordering of netlists without any modifications in the connectivity of the circuits.

\vspace{0.1cm}
Initially, the self-prepared dataset contains 27,092 (22,045 from Github and 5,047 from LTSpice XVII) publicly available .asc circuit files. After filtering and removing the training samples overlapping with the test samples, it is reduced to 10,428 different circuit schematics. This filtered dataset is split into 9,907 (95\%) training and 521 (5\%) validation circuit files, which after augmenting each circuit file into up to 5 samples, result in a total of 44,995 and 2,407 samples for training and validation, respectively. For testing, we obtain 117 filtered circuits out of 122 circuit examples from {\em Circuits-LTSpice}.

\subsection{Inference and fine-tuning}
We start with Llama 3.1 base \& instruct versions with 8 billion parameters (\llama{}) \cite{dubey2024llama3herdmodels} to develop \schemato{}. The tasks are first executed on the raw models to select the best performing prompts that are then used for fine-tuning. We use the recently proposed Torchtune~\cite{torchtune} framework for LLM fine-tuning. We first discuss the details of inference and fine-tuning of the models.
 
 During inference, \schemato{} generates schematics in an
auto-regressive way. Here, we used no special techniques to save
memory or compute except for KV-caches. We apply greedy search, where the LLM selects the most probable next sample. Decoding terminates either if maximum number of 
tokens ($N_{\text{tokens}}$) are reached or the stop-token is sampled. We choose $N_{\text{tokens}} =
8192$ as this is the maximum number of tokens that can fit in a single
NVIDIA Ada-6000 GPU. 
With this setup, the inference on \schemato{} achieved approximately $23$ tokens/sec on a single NVIDIA Ada-6000.

To adapt the LLM to our specific task of translating netlists into schematics, we employed supervised instruction tuning on a paired dataset
containing prompts with netlists and corresponding reference schematics as responses. 
However, fine-tuning LLMs like \schemato{} with 8 billion parameters requires a significant amount of GPU memory. To address this issue, we utilized
fully-sharded data parallel (FSDP) training \cite{zhao2023pytorchfsdpexperiencesscaling}, which enables the distribution of the model's parameters across multi-GPUs while minimizing communication overhead.
To further reduce memory consumption during training, we applied several additional techniques:
\begin{itemize}
    \item We used bfloat16 representation for model parameters to decrease the memory usage compared to float32.
    \item Activation checkpointing was employed to compute gradients in a more memory-efficient manner.
    \item Low-rank adaptation (LoRA) with rank $r=8$ and scaling factor $\alpha=16$ was applied to reduce the memory requirements of the trainable weights.
    \item Cross entropy with chunked output loss \cite{torchtune} is used to measure the difference between the generated token and the ground truth.
\end{itemize}

By combining these techniques, we fine-tune \llama{} over 10 epochs (in each epoch, \llama{} goes through the entire training set). The number of weight updates per epoch is:
\begin{equation*}
\frac{N_{\text{train}}}{N_{\text{iter}} \times N_{\text{GPU}} \times N_{\text{batch}}} = \frac{44,995}{8 \times 8 \times 1} \approx 703
\end{equation*}
where $N_{\text{train}}$ is the number of training samples, $N_{\text{iter}}$ is the number of iterations per accumulation step, $N_{\text{GPU}}$ is the number of NVIDIA Ada-6000 GPUs used for fine-tuning, and $N_{\text{batch}}$ is the per-GPU batch size. 
This setup achieved a throughput of approximately $563$ tokens/sec per GPU, allowing us to efficiently adapt the model to our specific task within a reasonable timeframe. We next discuss the different metrics used to evaluate the performance of the models. 

\subsection{Performance metrics} \label{Ch:metrics}
We use the following four metrics to evaluate the samples generated by the different models in order to capture various aspects of schematic generation: (1) graph edit distance (GED) score that measures the similarity of ground truth and generated graphs to compare their connectivity, (2) mean structural similarity index measure (MSSIM) that measures visual similarity with human generated design, (3) compilation success rate (CSR) that measures syntatic correctness, and (4) bilingual evaluation understudy (BLEU)  that measures an N-gram similarity. We describe these metrics next.

\subsubsection{Graph-based metric}
The graph edit distance (GED) measures the minimum number of modifications (inserting/deleting/substituting a node/edge) required to transform one graph to another. The algorithm to compute this metric is often implemented with shortest-path algorithms like the Dijkstra's algorithm \cite{dijkstra_based_ged}. As this metric is provided as an integer value, we need to normalize it such that it is agnostic to the graph size. We formulate the GED score as

\begin{equation*}
    \text{GED score} = 1- \frac{\text{GED}}{\text{max}(N_{\text{n, gen}}, N_{\text{n, ref}}) + \text{max}(N_{\text{e, gen}}, N_{\text{e, ref}})}
\end{equation*}

where $N_{\text{a, b}}$ is the number of a (n for nodes and e for edges) in b (gen for generated and ref for reference) samples.
The computation of $\text{GED}$ relies on the function, $\text{``graph\_edit\_distance"}$ from NetworkX \cite{SciPyProceedings_11}, which utilizes the depth-first graph edit distance algorithm \cite{abuaisheh:hal-01168816}. The time-out is set to 60 seconds for each sample pair comparison to prevent an excessively long search. The function returns the minimum GED found so far once this limit is reached. To map the generated asc files into the form of graphs, we again use LTSpice to compile them back into netlists. Then, we convert these netlists and the reference netlists provided in the prompt into nxgraph class using NetworkX and compute the GED scores. For cases where the generated asc script does not successfully compile into a netlist, we omit the sample from the GED score calculation. We then compare the performance of the models using $\overline{\mbox{GED score}}$, that is the GED score averaged over all compilable test samples.

\subsubsection{Image-based metric} 
The structural similarity index measure (SSIM) is a metric that computes the structural similarity of two images based on the statistical comparisons in luminance, contrast, and structure \cite{ssim}.
For a given pair of image patches $x$ and $y$, SSIM is defined as 
\begin{equation*}
\text{SSIM}(x, y) = \frac{(2\mu_x\mu_y + C_1)(2\sigma_{xy} + C_2)}{(\mu_x^2 + \mu_y^2 + C_1)(\sigma_x^2 + \sigma_y^2 + C_2)}
\end{equation*}
where $\mu_{p}$ and $\sigma_{p}$ are the average and the standard deviation of an image patch $p$, respectively,
$\sigma_{xy}$ is the square-root of the cross-correlation between the two image patches $x$ and $y$,
$C_1$ and $C_2$ are constants added to avoid computational instability.
To evaluate the overall similarity between an image pair $X$ and $Y$, a sliding local window is applied over the entire images and SSIM is computed at each pixel step within this window. The mean SSIM (MSSIM) between two images $X$ and $Y$ is then given by 
\begin{equation*}
    \text{MSSIM}(X, Y) = \frac{1}{M} \sum\nolimits_{i=1}^{M} \text{SSIM}(x_i, y_i)
\end{equation*}
where $x_i$ and $y_i$ are the image patches in the $i$-th local window and $M$ is the number of local windows applied on the images.
We use this metric to derive the similarities between the reference schematics in the test set and the schematics generated by LLMs from netlists. To obtain the images of generated and reference schematics, we load each .asc file with LTSpice and take a screenshot. These images are subsequently cropped to eliminate the artifacts from the graphical user interface. 
Similarly to the GED scores, we omit the incompilable samples from the calculation. We then compare the performance of the models using $\overline{\mbox{MSSIM}}$, the MSSIM score averaged over all compilable test samples.

\subsubsection{Compilation-based metric}
The compilation success rate (CSR) measures the syntactical correctness of the generated .asc files and is defined as
\begin{equation*}
    \mbox{CSR}=\displaystyle\frac{N_\text{compilable}}{N_{\text{ref}}}
\end{equation*}
where $N_\text{compilable}$ is the total number of generated schematics in the test set that can be compiled into netlists using LTSpice and $N_{\text{ref}}$ is the total number of reference samples.

\subsubsection{Text-based metric}
The bilingual evaluation understudy (BLEU) compares N-grams of the generated text and the reference translation and counts the number of matches independently of the positions \cite{Papineni2002}. With this measure, we compare the N-gram similarity between the reference and generated text-based schematics. While BLEU is not perfectly suited for our tasks due to the syntactical nature of .asc files, it can indicate the extent to which the pre-trained models already understand their file syntax. 
Furthermore, we selected the standard BLEU as our metric over CodeBLEU \cite{ren2020codebleumethodautomaticevaluation} specialized to capture syntactic and semantic features of codes since this measure does not support .asc file formats.
We compare the performance of the models by computing $\overline{\text{BLEU}}$, that is the average BLEU score across all test samples.


\begin{table}[t]
\scriptsize
    \centering
    \caption{Test performance comparison. $\overline{\text{GED}}$ is equivalent to the average GED score, $\overline{\text{GED score}}$. The values on the left and right side of / indicate the raw scores and scores scaled by CSR, respectively for $\overline{\text{GED score}}$ and $\overline{\text{MSSIM}}$. The best and second best metrics are highlighted in {\textcolor{red}{red}} and {\textcolor{blue}{blue}} respectively.}
    \label{Table:net-asc1}
   \begin{tabular}{>{\centering\arraybackslash}p{1.4cm} >{\centering\arraybackslash}p{0.5cm} >{\centering\arraybackslash}p{0.5cm} >{\centering\arraybackslash}p{0.7cm} >{\centering\arraybackslash}p{0.8cm} >{\centering\arraybackslash}p{0.6cm} >{\centering\arraybackslash}p{0.6cm}}
    \toprule
     \textbf{Model} & \textbf{Version} & \textbf{Prompt} & $\mathbf{\overline{GED}}$ & $\mathbf{\overline{MSSIM}}$ & \textbf{CSR}(\%) & $\mathbf{\overline{BLEU}}$ \\
    
    \midrule
    \rowcol
    \multicolumn{7}{c}{\emph{Pretrained models}} \\
 Llama-3.1-8B & base & 1 & - & - & 0.00 & 0.00\\
 Llama-3.1-8B & instruct & 1 & - & - & 0.00 & 0.00\\
 GPT & 4o & 1 & 0.21/0.06 & 0.05/0.01 & 27.35 & 9.24\\
 
 Llama-3.1-8B & base & 2 & 0.26/0.01 & 0.02/0.00 & 3.42 & 6.40\\
 Llama-3.1-8B & instruct & 2 & 0.26/0.01 & 0.02/0.00 & 3.42 & 6.40\\
 GPT & 4o & 2 & 0.22/0.06 & 0.04/0.01 & 27.35 & 11.97\\

 Llama-3.1-8B & base & 3 & - & - & 0.00 & 0.00\\
 Llama-3.1-8B & instruct & 3 & - & - & 0.00 & 0.00\\
 GPT & 4o & 3 & 0.13/0.00 & 0.02/0.00 & 0.85 & 8.76\\

 Llama-3.1-8B & base & 4 & \textcolor{blue}{0.30}/0.13 & 0.06/0.03 & 41.88 & 10.51\\
 Llama-3.1-8B & instruct & 4 & \textcolor{blue}{0.30}/0.13 & 0.03/0.01 & 41.88 & 10.51\\
 GPT & 4o & 4 & 0.23/0.12 & \textcolor{blue}{0.07}/\textcolor{blue}{0.04} & 53.85 & \textcolor{blue}{15.22}\\

 Llama-3.1-8B & base & 5 & 0.27/0.10 & 0.06/0.02 & 38.46 & 10.55\\
 Llama-3.1-8B & instruct & 5 & 0.27/0.10 & 0.06/0.02 & 38.46 & 10.55\\
 GPT & 4o & 5 & 0.23/\textcolor{blue}{0.15} & 0.03/0.02 & \textcolor{blue}{63.25} & 14.84\\

    \midrule
    \rowcol
    \multicolumn{7}{c}{\emph{Best fine-tuned model}} \\ 
    \schemato{} & instruct & 3 & \textcolor{red}{0.35}/\textcolor{red}{0.27} & \textcolor{red}{0.22}/\textcolor{red}{0.17} & \textcolor{red}{76.07} & \textcolor{red}{22.17}\\
    
    \bottomrule
    \end{tabular}
\end{table}

\begin{table}[t]
\scriptsize
    \centering
    \caption{Ablation study for base model selected for fine-tuning. The values on the left and right side of / indicate the raw scores and scores scaled by CSR, respectively for $\overline{\text{GED score}}$ and $\overline{\text{MSSIM}}$. For each prompt, the best performing weights are chosen. The best metrics are highlighted in {\textcolor{red}{red}}.}
    \label{Table:net-asc2}
   \begin{tabular}{>{\centering\arraybackslash}p{1.4cm} > {\centering\arraybackslash}p{0.4cm} >{\centering\arraybackslash}p{0.5cm} >{\centering\arraybackslash}p{0.5cm} >{\centering\arraybackslash}p{0.7cm} >{\centering\arraybackslash}p{0.8cm} >{\centering\arraybackslash}p{0.6cm} >{\centering\arraybackslash}p{0.6cm}}
    \toprule
     \textbf{Model} & \textbf{Epoch} & \textbf{Version} & \textbf{Prompt} & $\mathbf{\overline{GED}}$ & $\mathbf{\overline{MSSIM}}$ & \textbf{CSR}(\%) & $\mathbf{\overline{BLEU}}$ \\
   \midrule
    \rowcol
    \multicolumn{8}{c}{\emph{Models fine-tuned with different prompts}}\\
    
   \llama{} & 4 & instruct & 2 & 0.39/0.27 & 0.03/0.02 & 70.09 & 20.11\\
  \llama{} & 5 & instruct & 4 & \textcolor{red}{0.43}/\textcolor{red}{0.28} & 0.04/0.03 & 64.10 & 15.26\\
 \llama{} & 9 & instruct & 5 & 0.33/0.26 & 0.20/0.16 & \textcolor{red}{77.78} & 
 \textcolor{red}{23.40}\\

    \midrule
    \rowcol
    \multicolumn{8}{c}{\emph{Best fine-tuned model}} \\ 
    \schemato{} & 6 & instruct & 3 & 0.35/0.27 & \textcolor{red}{0.22}/\textcolor{red}{0.17} & 76.07 & 22.17\\
    
    \bottomrule
    \end{tabular}
\end{table}

\begin{figure}[t]
    \centering
    \includegraphics[width=0.48\textwidth,trim=10 0 25 10,clip]{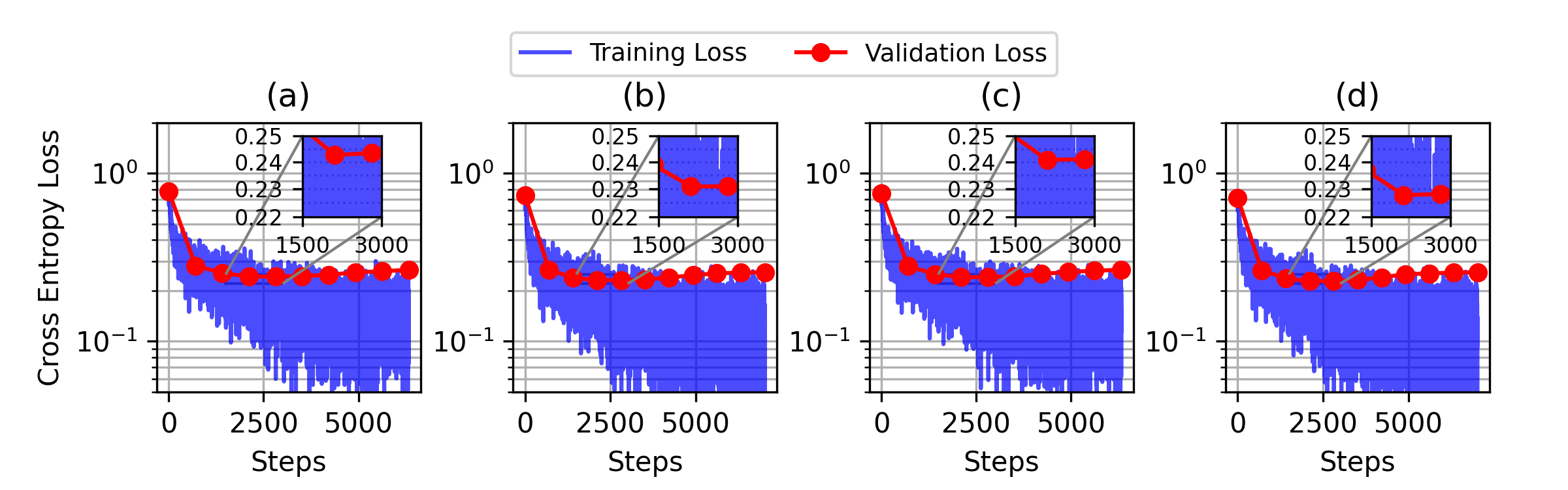}
    \caption{Training and validation losses. \llama{} is fine-tuned with four different prompts: (a) \mbox{Prompt 2}, (b) \mbox{Prompt 3}, (c) \mbox{Prompt 4}, and (d) \mbox{Prompt 5}. Validations losses are computed after every training epoch.}
    \label{fig:loss}
    \vspace{-0.3cm}
\end{figure}

\section{Results \& Discussion} \label{Ch:result}
In order to select the starting model for fine-tuning, we evaluated the performance of \llama{} raw models across various configurations. \autoref{Table:net-asc1} shows the performance of the different models on the evaluation using the test set. 
As can be seen, the performance did not vary across the two versions: base and instruct. Moving from zero-shot (\mbox{Prompt 1-3}) to one-shot prompting (\mbox{Prompt 4-5}) improved the best CSR significantly from 3.42 to 41.88\% while its impact is difficult to observe for GED scores and MSSIM due to the low number of compiled samples (i.e. low CSR) for zero-shot prompting. The effect of providing the schematic sheet size is negligible across all the metrics between \mbox{Prompt 4} to \mbox{Prompt 5}. 
We select \mbox{Prompt 2} for fine-tuning as it performs well out of all the zero-shot prompts and \mbox{Prompt 4} due to the observed superiority in CSR. While, the effect of providing sheet size is not observable from this result, we also include \mbox{Prompt 3} and 5 to examine how this affects the performance of the fine-tuned model. Furthermore, we used \llama{} instruct model to leverage its knowledge base for text synthesis.

\begin{figure}[t]
    \centering
    \includegraphics[width=0.48\textwidth,trim=15 0 25 0,clip]{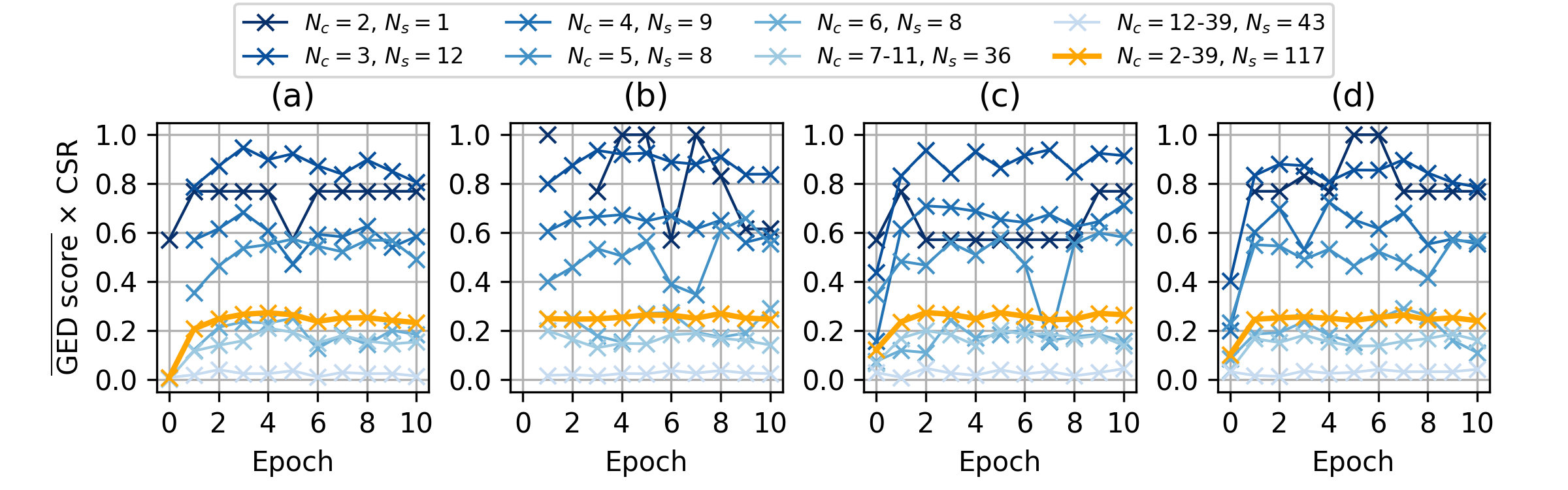}
    
    \caption{$\overline{\text{GED score}}$s scaled by CSR evaluated with the test set. \llama{} is fine-tuned with (a) \mbox{Prompt 2}, (b) \mbox{Prompt 3}, (c) \mbox{Prompt 4}, and (d) \mbox{Prompt 5}. The missing data points indicate that no generated samples are compilable (i.e. $\text{CSR}=0$). The number of components in the test circuits is denoted by $N_c$ and the number of reference samples composed of $N_c$ components is denoted by $N_{s}$.}
    \label{fig:CSR_GED}
    \vspace{-0.3cm}    
\end{figure}

\begin{figure}[t]
    \centering
    \includegraphics[width=0.48\textwidth,trim=10 0 25 0,clip]{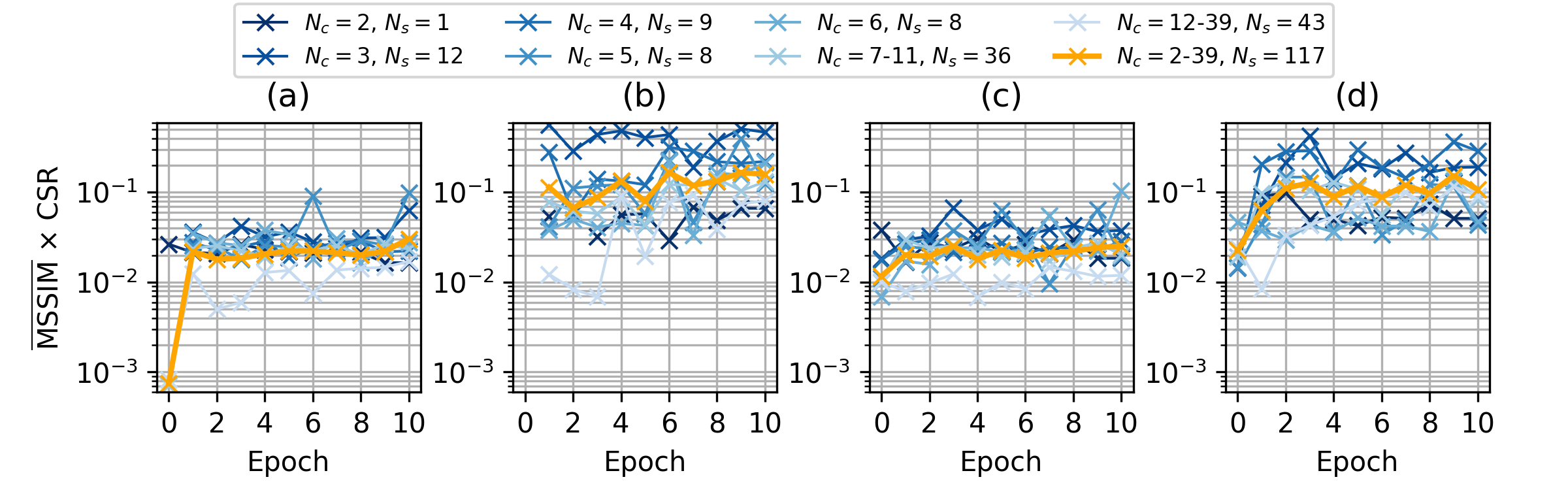}
    
    \caption{$\overline{\text{MSSIM}}$ scaled by CSR evaluated with the test set. \llama{} is fine-tuned with (a) \mbox{Prompt 2}, (b) \mbox{Prompt 3}, (c) \mbox{Prompt 4}, and (d) \mbox{Prompt 5}. The missing data points indicate that no generated samples are compilable (i.e. $\text{CSR}=0$). The number of components in the test circuits is denoted by $N_c$ and the number of reference samples composed of $N_c$ components is denoted by $N_{s}$.}
    \label{fig:CSR_MSSIM}
    \vspace{-0.2cm}    
\end{figure}

During fine-tuning, training and validation losses (\autoref{fig:loss}) revealed overfitting (an increase in validation loss) after epoch 3. 
Since the cross entropy loss does not directly help quantify the correctness of the code and the generated schematics, we use $\overline{\text{GED score}}$ and $\overline{\text{MSSIM}}$ discussed in Section \ref{Ch:metrics} to evaluate the model performance across training epochs. $\overline{\text{GED score}}$s scaled by CSR for samples containing different numbers of components from epoch 0 (raw model of \llama) to epoch 10 are shown in \autoref{fig:CSR_GED}. We scale the scores by CSR for a fairer comparison since samples that are more difficult to generate (e.g. circuits with a large number of components) can become compilable during fine-tuning and be included in the calculation of the average score. This can prevent the scores from improving because these larger circuits tend to have lower scores than the smaller circuits as observed. 
The effect of different prompting techniques is not observable while the fine-tuning significantly improves the performance until the model overfits around epoch 3, which aligns with the observation from \autoref{fig:loss}.
Similarly, $\overline{\text{MSSIM}}$ scaled by CSR for different circuit size is shown in \autoref{fig:CSR_MSSIM}. In contrast to the previous observations, the scores improve until the last epoch. Furthermore, the impact of providing schematic sheet size is clearly reflected by the scores that are almost one-order of magnitude higher for (b) \mbox{Prompt 3} and (d) \mbox{Prompt 5} while one-shot prompting did not impact the performance.

We then choose \llama{} fine-tuned with \mbox{Prompt 3} using weights from epoch 6 as our best model and call it \schemato{} since it demonstrated the best combination of scores in \autoref{fig:CSR_GED} and \autoref{fig:CSR_MSSIM}. We evaluate this model by comparing its performance with the \llama{} baselines and GPT-4o, the latest and most powerful model from OpenAI. 
\autoref{Table:net-asc1} shows that \schemato{} outperforms the other models with $1.8\times$ and $4.3\times$ higher $\overline{\text{GED score}}$ and $\overline{\text{MSSIM}}$ scaled by CSR than the best performing pretrained models. Furthermore, \schemato{} scores $76\%$ CSR compared to $63\%$ scored by the best performing pretrained model. \autoref{Table:net-asc2} shows the ablation study of \llama{} instruct fine-tuned with different prompts.
There is no significant difference in the scores between the models fine-tuned with the zero-shot and one-shot prompts. In contrast, the dominance of the models fine-tuned with \mbox{Prompt 3} (\schemato{}) and \mbox{Prompt 5} in $\overline{\text{MSSIM}}$ and CSR showcases the effectiveness of including sheet size in the prompt.

\begin{figure}[t]
\vspace{-.1in}
\hspace{.1in}
    \centering
    \begin{subfigure}{0.25\textwidth} 
        \centering
        \raisebox{0.25cm}{
        \begin{tcolorbox}[colback=gray!5!white, colframe=gray!75!black]
            \scriptsize
            T1 N002 0 N003 0 Td=50n Z0=50 \\
            V1 N001 0 V \\
            RS N002 N001 R \\
            RL N003 0 R 
        \end{tcolorbox}}
        \caption{Input netlist}
    \end{subfigure}%
    \begin{subfigure}{0.23\textwidth} 
        \centering
            \includegraphics[height=2.4cm,trim=10 200 10 200,clip]{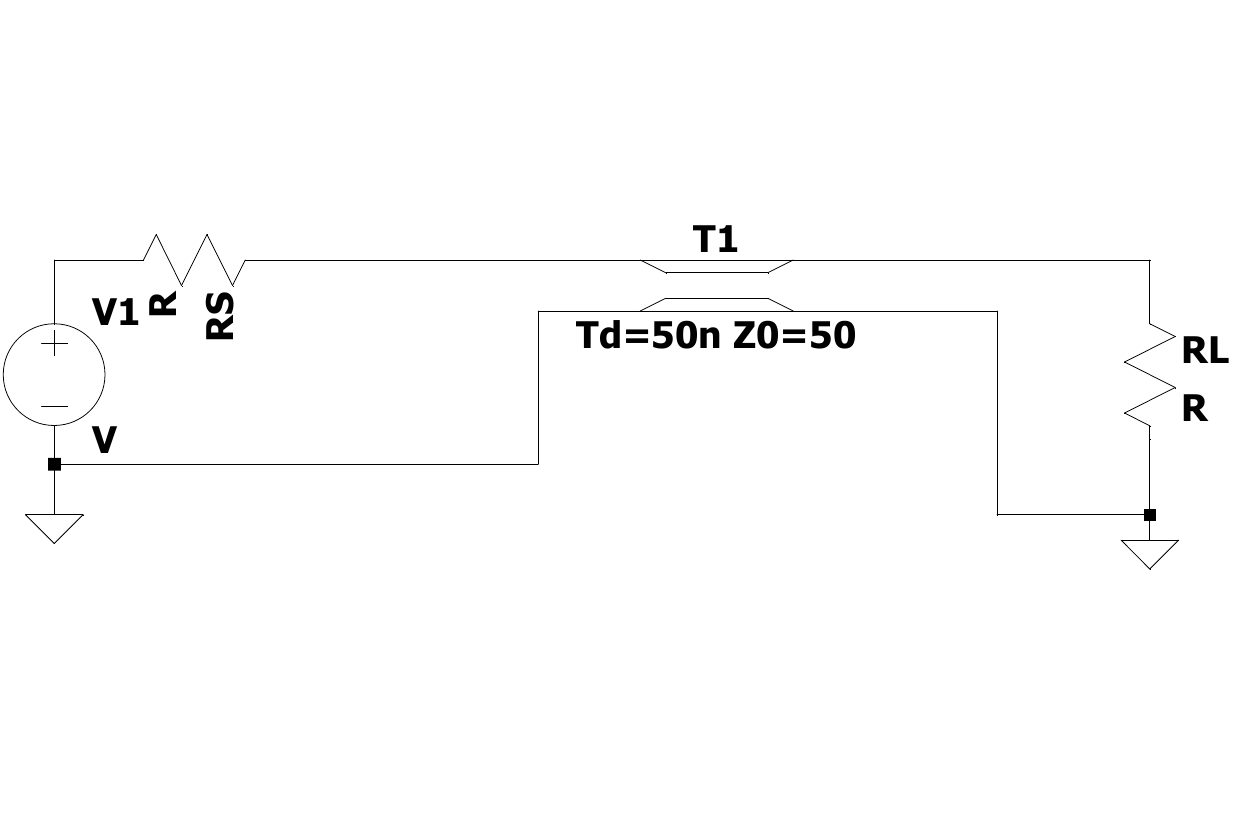}
            \caption{Reference schematic}
    \end{subfigure}%
    \hfill
    \begin{subfigure}{0.25\textwidth} 
        \centering
            \includegraphics[height=2.5cm,trim=5 200 5 150,clip]{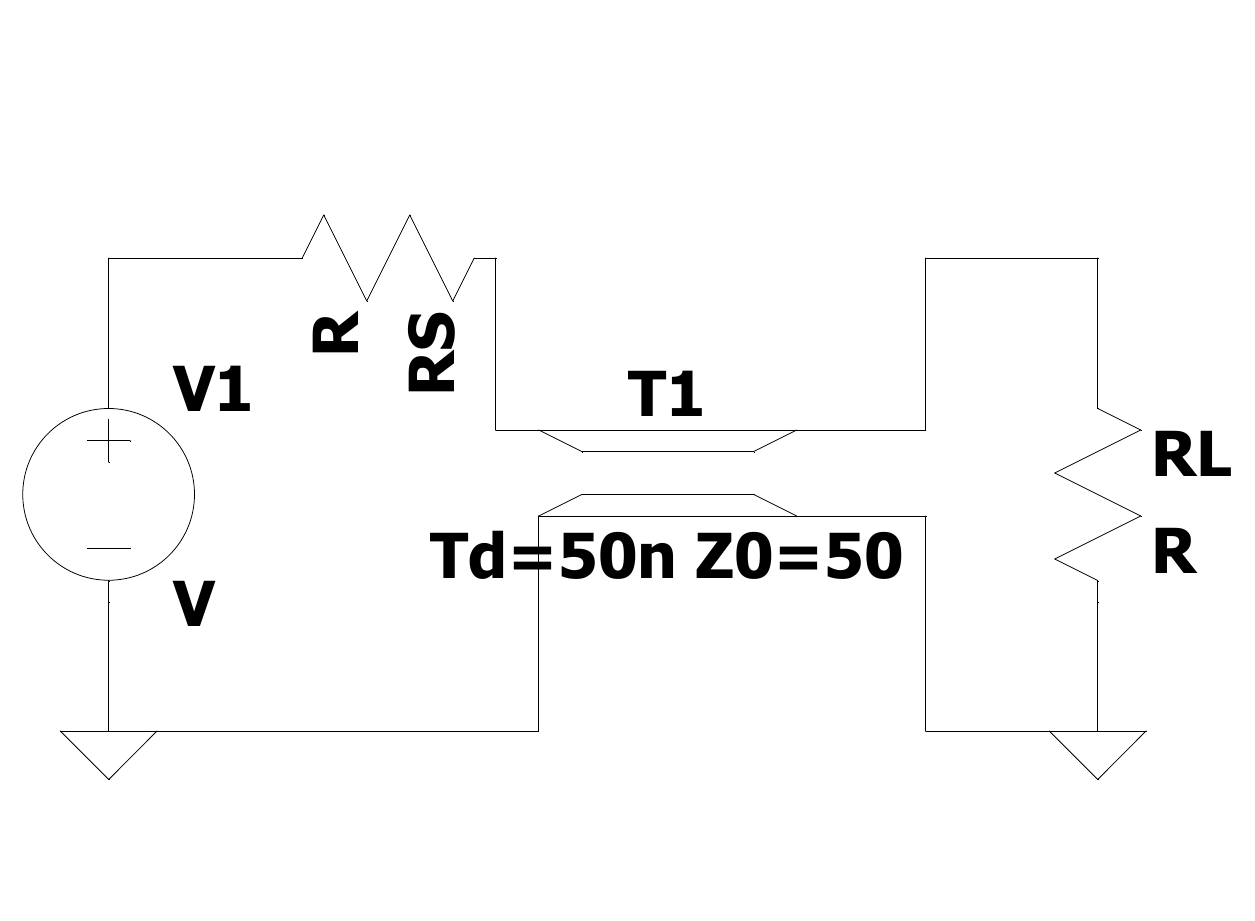}
            \captionsetup{justification=centering} 
            \caption{\schemato{}}
    \end{subfigure}%
    \hfill
    \begin{subfigure}{0.23\textwidth}
        \centering
            \includegraphics[height=2.3cm,trim=20 20 20 0,clip]{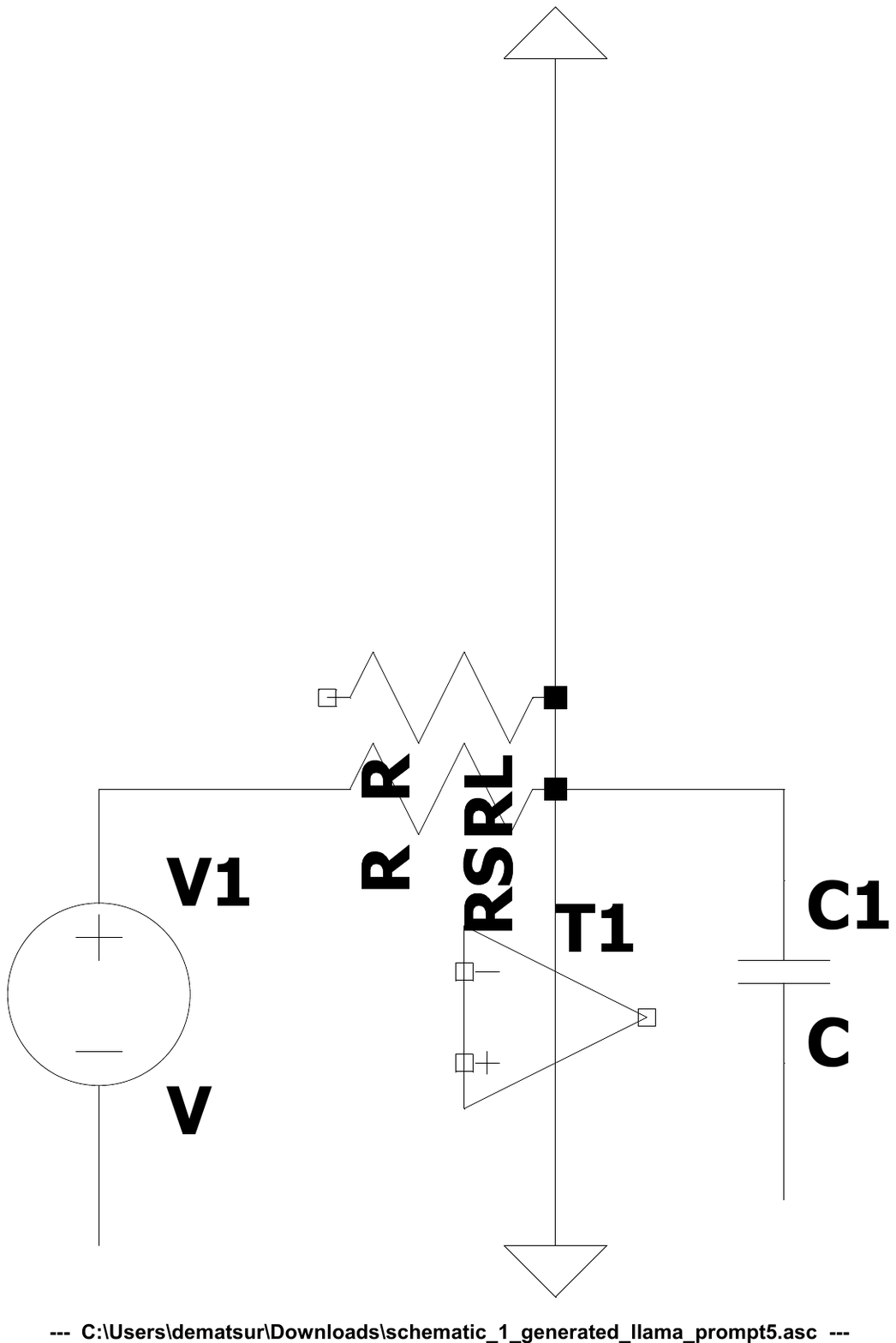}
            \captionsetup{justification=centering} 
            \caption{Llama 3.1 instruct \\with \mbox{Prompt 5}}
    \end{subfigure}%

    \caption{Netlist \& LTSpice schematics of a transformer circuit}
    \label{fig:schematics_ltspice1}
    \vspace{-0.3cm}
\end{figure}

\begin{figure}[t]
\vspace{-.1in}
\hspace{.1in}
    \centering
    \begin{subfigure}{0.18\textwidth} 
        \centering
        \raisebox{0.25cm}{
        \begin{tcolorbox}[colback=gray!5!white, colframe=gray!75!black]
            \scriptsize
            V1 N001 0 V\\
            C1 N002 N001 C\\
            R1 N002 Vout R\\
            R2 Vout 0 R\\
            C2 Vout 0 C
        \end{tcolorbox}}
        \caption{Input netlist}
    \end{subfigure}%
    \begin{subfigure}{0.33\textwidth} 
        \centering
            \includegraphics[height=2.7cm,trim=10 200 10 200,clip]{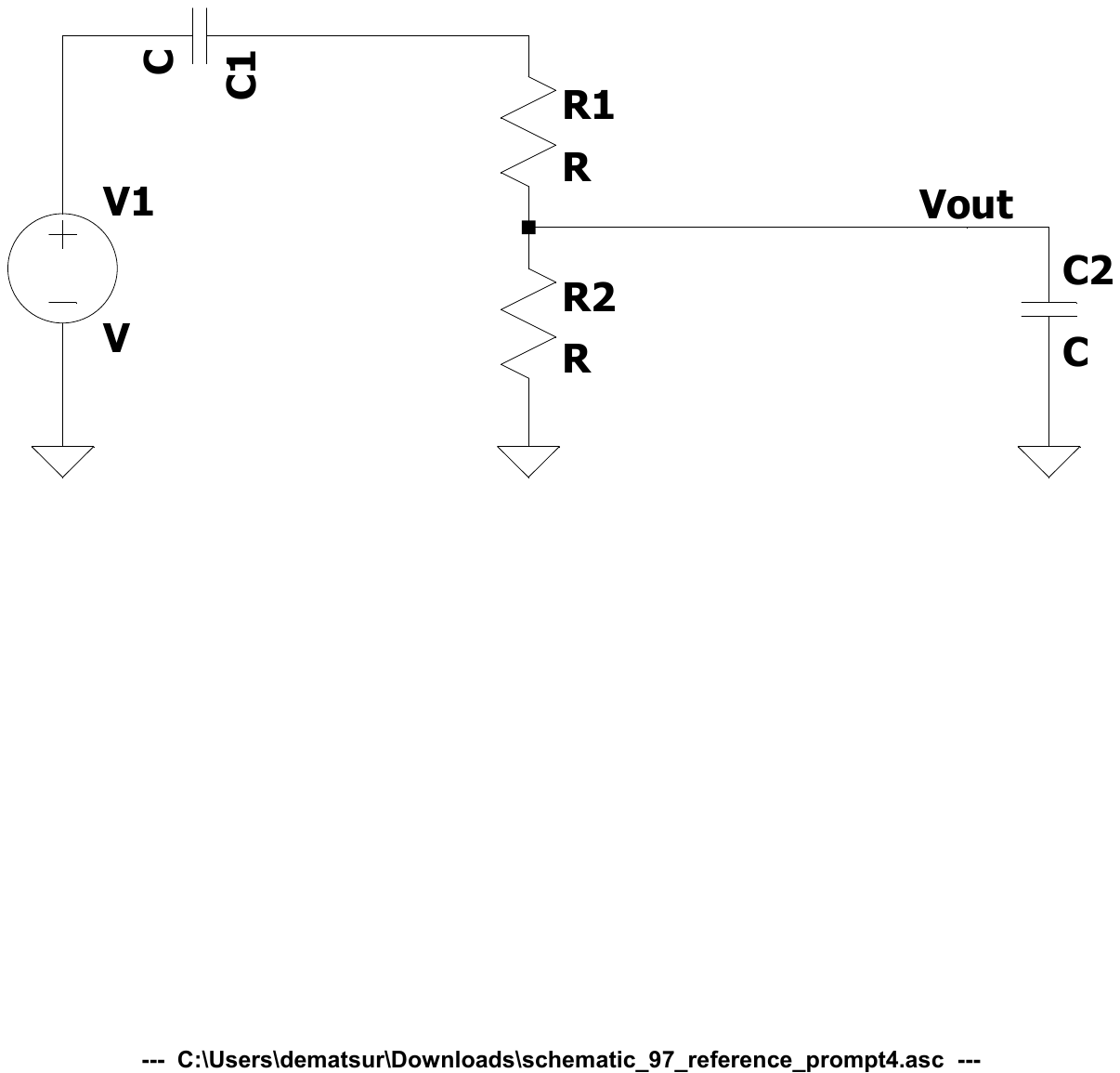}
            \caption{Reference schematic}
    \end{subfigure}%
    \hfill
    \begin{subfigure}{0.18\textwidth} 
        \centering
            \includegraphics[height=2.5cm,trim=5 200 5 150,clip]{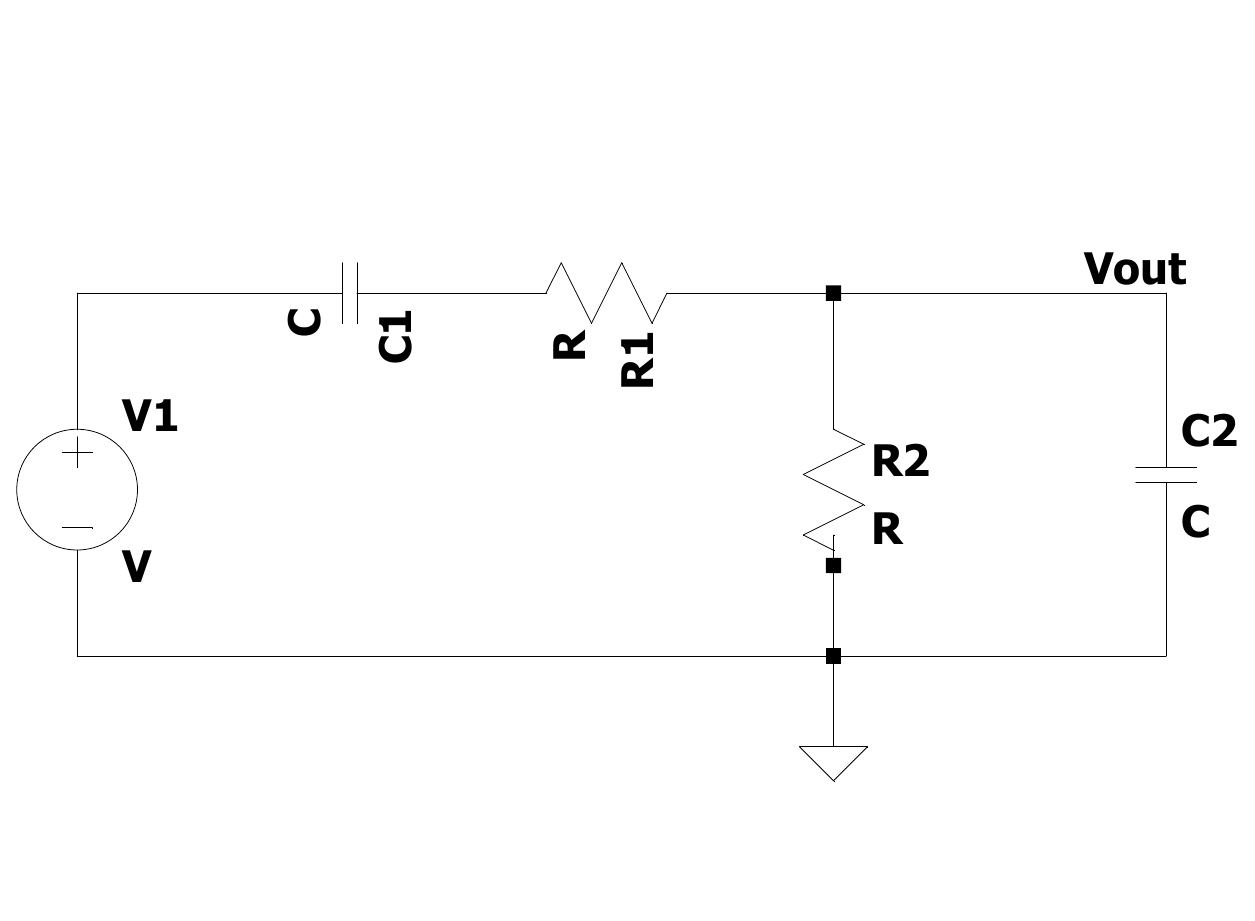}
            \captionsetup{justification=centering} 
            \caption{\schemato{}}
    \end{subfigure}%
    \hfill
    \begin{subfigure}{0.15\textwidth}
        \centering
            \includegraphics[height=2.3cm,trim=20 20 20 0,clip]{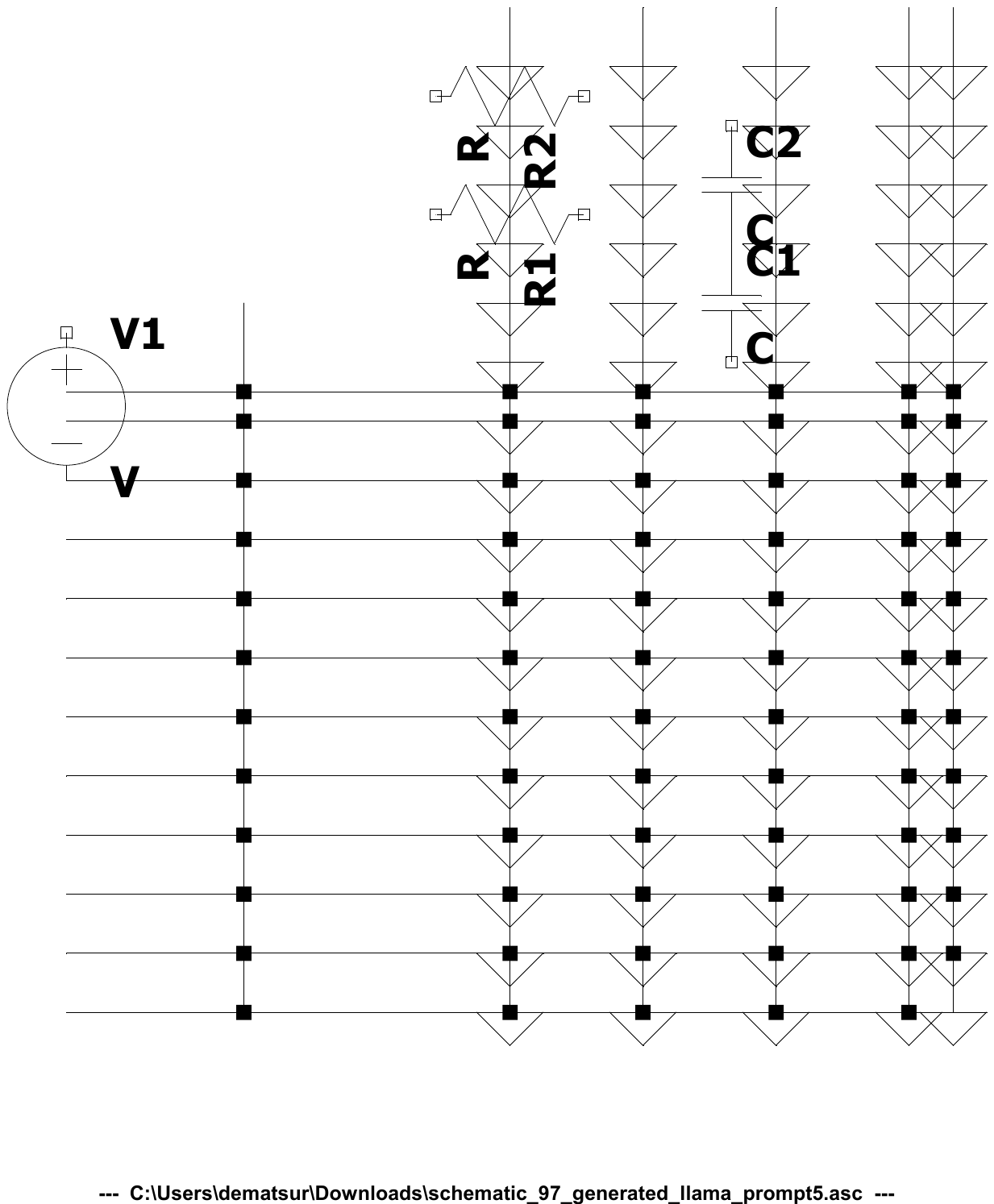}
            \captionsetup{justification=centering} 
            \caption{Llama 3.1 instruct \\with \mbox{Prompt 5}}
    \end{subfigure}%
    \hfill
    \begin{subfigure}{0.15\textwidth}
        \centering
        \includegraphics[height=2.2cm,trim=0 170 0 80,clip]{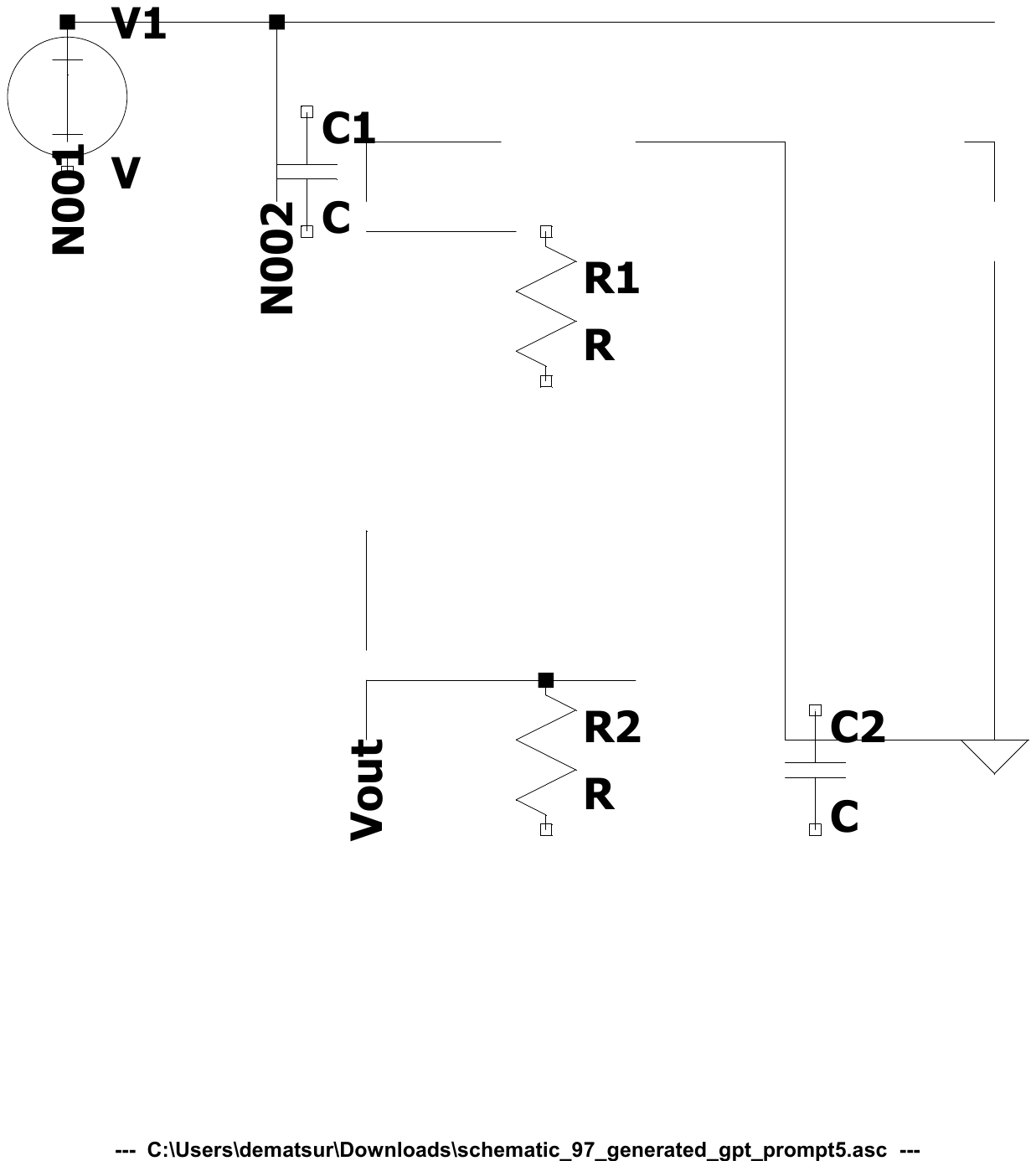}
        \captionsetup{justification=centering} 
            \caption{GPT-4o \\with \mbox{Prompt 5}}
    \end{subfigure}%

    \caption{Netlist \& LTSpice schematics of an RC band-pass filter}
    \label{fig:schematics_ltspice2}
    \vspace{-0.3cm}
\end{figure}

We show the reference and generated schematics for test circuit examples of a transformer circuit and an RC band-pass filter in \autoref{fig:schematics_ltspice1} and \autoref{fig:schematics_ltspice2} for different models. We note that GPT-4o failed to produce a compilable code for the first example across all the prompts. 
Although \schemato{}'s schematics differ geometrically from the reference ($\text{MSSIM} \neq 1$), they maintain identical topological features ($\text{GED score}=1$), demonstrating successful netlist-to-schematic conversion and \schemato{}'s ability to generalize over different circuit topologies.

\section{Conclusions \& Future work}
\label{sec:conclusions}
In this paper, we introduced and evaluated \schemato{}, a fine-tuned LLM designed for converting circuit netlists into interpretable visual schematics. Our experimental results demonstrate that \schemato{} significantly outperforms state-of-the-art LLMs, including \llama{} and GPT-4o, across all the graph-based, image-based, compilation-based, and text-based metrics. These findings underscore the potential of \schemato{} in reliably generating human-intuitive schematics from netlists, paving the way for enhanced interpretability and utility in ML-based analog circuit synthesis.

Our experiment in \autoref{fig:CSR_GED} show that for circuits with more than 5 components, \schemato{} struggles to generate schematics with accurate connectivity.  
This is expected as there are more possible variations of circuits with different connectivity and components for these larger circuits, making it harder for \schemato{} to generalize. 
The .asc file format further complicates generalization capabilities since it only specifies the components' name and coordinate, but not symbol size. This means that \schemato{} has to learn their symbol size from how wires are placed around the components across our training samples and therefore the model cannot generate schematics containing less familiar or unfamiliar components. The analysis on our dataset revealed that 5,227 out of the 9,907 training samples contain components that appear less than 10 times in our training set. Hence, more than half is composed of components of low familiarity.
This highlights the need for a larger dataset with circuits composed of more generic and constrained set of components that are used in various configurations.

To address this challenge, future improvements to \schemato{} will focus on enriching the dataset using advanced data augmentation techniques. For instance, we could combine two schematics into a new one using a technique similar to mixup used in vision~\cite{zhang2018mixupempiricalriskminimization}. To improve the generalizability of {\schemato} to large and unseen circuit topologies, we will investigate the decomposition of larger circuits into smaller, more manageable subcircuits, thereby enhancing the ability of the model to learn hierarchical structures and the interplay between subcircuits. In addition, we plan to further expand the dataset by converting widely available schematic images into netlists and digitized schematics, using a method similar to that implemented in \cite{shi2025amsnet20largeams}. 


\newpage
\IEEEtriggeratref{16}
\bibliographystyle{IEEEtran}
\bibliography{references_local}
\end{document}